\documentclass[10pt,twocolumn,letterpaper]{article}

\usepackage[accsupp]{axessibility}  
\usepackage{iccv}
\usepackage{times}
\usepackage{epsfig}
\usepackage{graphicx}
\usepackage{amsmath}
\usepackage{amssymb}

\usepackage[pagebackref=true,breaklinks=true,letterpaper=true,colorlinks,bookmarks=false]{hyperref}

\iccvfinalcopy 

\usepackage[capitalize]{cleveref}
\crefname{section}{Sec.}{Secs.}
\Crefname{section}{Section}{Sections}
\Crefname{table}{Table}{Tables}
\crefname{table}{Tab.}{Tabs.}
\usepackage{multirow}
\usepackage{booktabs}
\makeatletter
\newcommand\dlmu[2][4cm]{\hskip1pt\underline{\hb@xt@ #1{\hss#2\hss}}\hskip3pt}
\makeatother

\def\iccvPaperID{7398} 

\ificcvfinal\fi

\begin{document}

\title{BEVPlace: Learning LiDAR-based Place Recognition using Bird's Eye View Images}


	\author{Lun Luo$^{1,2,4}$, Shuhang Zheng$^{2}$, Yixuan Li$^{2}$, Yongzhi Fan$^{2}$, Beinan Yu$^{2}$, Si-Yuan Cao$^{1,2*}$,  \\ Junwei Li$^{1,2}$, Hui-Liang Shen$^{1,2,3}$\thanks{Corresponding author.}\\
	$^{1}\text{Ningbo Innovation Center, Zhejiang University}$ \\ 	
	$^{2} \text{College of Information Science and Electronic Engineering, Zhejiang University	}$\\
    $^{3}\text{Key Laboratory of Collaborative Sensing and Autonomous Unmanned Systems of Zhejiang Province}$ \\
	$^{4}\text{HAOMO.AI Technology Co., Ltd.}$ \\
	{\tt\small \{luolun, zhengsh, yixuanli, tony\_fan, yubeinan, cao\_siyuan, lijunwei7788, shenhl\}@zju.edu.cn} \\
}

\maketitle
\ificcvfinal\thispagestyle{empty}\fi

\begin{abstract}
   Place recognition is a key module for long-term SLAM systems. Current LiDAR-based place recognition methods usually use representations of point clouds such as unordered points or range images. These methods achieve high recall rates of retrieval, but their performance may degrade in the case of view variation or scene changes. In this work, we explore the potential of a different representation in place recognition, i.e. bird's eye view (BEV) images. We validate that, in scenes of slight viewpoint changes, a simple NetVLAD network trained on BEV images achieves comparable performance to the state-of-the-art place recognition methods. For robustness to view variations, we propose a rotation-invariant network called BEVPlace. We use group convolution to extract rotation-equivariant local features from the images and NetVLAD for global feature aggregation. In addition, we observe that the distance between BEV features is correlated with the geometry distance of point clouds. Based on the observation, we develop a method to estimate the position of the query cloud, extending the usage of place recognition. The experiments conducted on large-scale public datasets show that our method 1) achieves state-of-the-art performance in terms of recall rates, 2) is robust to view changes, 3) shows strong generalization ability, and 4) can estimate the positions of query point clouds. Source codes are publicly available at  \href{https://github.com/zjuluolun/BEVPlace}{https://github.com/zjuluolun/BEVPlace} .
\end{abstract}

\begin{figure}[ht]
	\begin{center} 
		\includegraphics [width=2.7in]{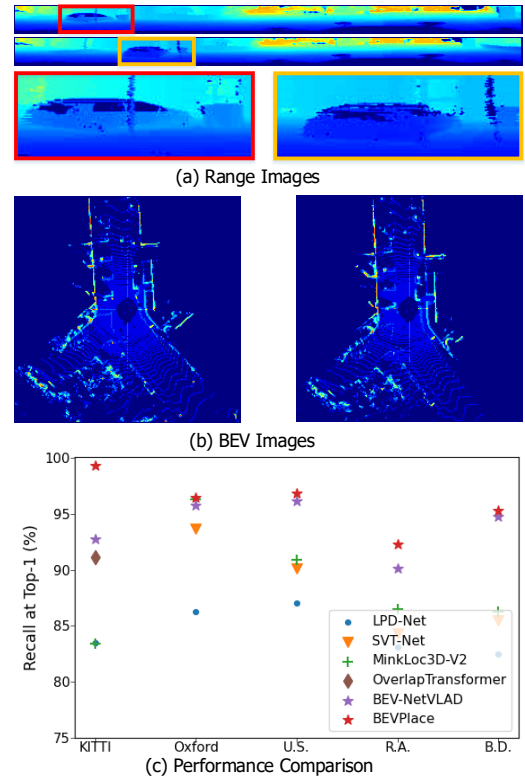}  
		\caption{(a) Two range images projected from two point clouds of KITTI that are 5 meters apart from each other. A small point cloud translation will cause distortions such as scale changes and occlusion. (b) The corresponding BEV images. The scale and position distribution of objects on the road almost remains unchanged. (c) Performance on various datasets. A simple BEV-based NetVALD network achieves comparable Top-1 recall to the SOTA methods. Our BEVPlace enhances the baseline further.}
		\label{fig: represention}
		\vspace{-10mm}
	\end{center} 

\end{figure}

\section{Introduction}
\label{sec:intro}
Place recognition plays an important role in both the map construction and localization phases of long-term Simultaneous Localization and Mapping (SLAM) systems \cite{cadena2016past}. In the map construction phase, it can provide loop closure constraints to eliminate the accumulated drift of the odometry. In the localization phase, it can re-localize the system when the pose tracking is lost and improve the robustness of the system. In recent years, lots of image-based place recognition methods \cite{2012Bags,arandjelovic2016netvlad,2017ORB} have been developed and achieved satisfactory performance. However, these methods are vulnerable to illumination changes and view variation due to the imaging mechanism of camera sensors. On the contrary, point clouds of LiDAR sensors are robust to illumination changes due to active sensing. In addition, the availability of precise depth information can help more accurate place recognition \cite{angelina2018pointnetvlad,liu2019lpd}.

LiDAR-based place recognition can be regarded as a retrieval problem, that is, finding the most similar frame to a query from a pre-built database. The key to solving this problem is to generate a global feature that can model the similarity between point clouds. PointNetVLAD \cite{angelina2018pointnetvlad} gives the first deep-learning solution to the problem of large-scale LiDAR-based place recognition. It uses PointNet \cite{qi2017pointnet} to extract local features from unordered points and NetVLAD \cite{arandjelovic2016netvlad} to generate global features. There are lots of subsequent methods that follow PointNetVLAD and introduce auxiliary modules such as attentions \cite{zhang2019pcan,sun2020dagc}, handcrafted features \cite{liu2019lpd}, and sparse convolution \cite{mickloc3d}. Recently, some methods \cite{chen2020overlapnet,OverlapTransformer} based on range images have been developed. The range image is the sphere projection of a point cloud. Due to the projection mechanism, the translation of the range image is equivariant to the rotation of the point cloud. Based on this, OverlapTransformer \cite{OverlapTransformer} uses a convolution network and a transformer to extract rotation-invariant features from the images. Some methods \cite{kim2018scan,kim20191,chen2020overlapnet} use similar projections and also achieve place recognition robust to view changes. 

Although the aforementioned methods have made great progress, they still have limitations in terms of generalization ability. This is because both unordered points and range images used for place recognition are sensitive to the motions of the LiDAR sensor. Specifically, for unordered points, the point coordinate and the relative positions between points will change severely along with motions of the LiDAR sensor. For range images, the image contents suffer various distortions with translations of point clouds although they are robust to rotations. Current methods \cite{angelina2018pointnetvlad,zhang2019pcan,liu2019lpd} force the network to learn these variations of data with data augmentation. However, as pointed out in \cite{tipooling}, data augmentation needs the network to be as flexible as possible to capture all the variations, which may result in the large risk of overfitting and poor generalization ability.

In this work, we explore the potential of place recognition using bird's eye view (BEV) images. The BEV image is generated by projecting a point cloud to the ground space. In road scenes, the transformations of point clouds are approximately equivariant to the rotations and translations of BEV images \cite{bvmatch2021}. Thus, the contents of BEV images are more robust to sensor motions. As shown in Fig. \ref{fig: represention}, the translation of a point cloud causes little appearance changes in the BEV image but introduces geometry distortions to the range image. The results shown in Fig. \ref{fig: represention} (c) validates that a simple NetVLAD based on the BEV representation achieves comparable performance with the state-of-the art methods.  To achieve robustness to viewpoint changes, we design a group convolution \cite{groupconv} network to extract local features from BEV images. Then, we use NetVLAD \cite{arandjelovic2016netvlad} for global rotation-invariant features extraction. Benefiting from the design of rotation invariance for BEV images, our method has the strong ability of place retrieval in the cases of both viewpoint variations and scene changes. In addition, we observe that the distances of the BEV features correlate well with the geometry distances of point clouds. According to this correlation, we map the feature distance to the geometry distance and then estimate the position of the query cloud, which extends the usage of LiDAR-based place recognition.

We summarize the contributions of this paper as follows:
\begin{itemize}
	\item We experimentally show that, without any delicate design, a simple NetVLAD network based on the BEV representation outperforms SOTA methods on the KITTI dataset\cite{kitti} and the benchmark dataset\cite{angelina2018pointnetvlad}.
	
	\item We propose a novel LiDAR-based place recognition method called BEVPlace. The method is robust to view changes, has strong generalization ability, and achieves SOTA performance on three large-scale datasets.
	
	\item We explore the statistical correlation between the feature distance and the geometry distance of point cloud pairs. To the best of our knowledge, this paper is the first to perform position estimation directly from global descriptors. 
\end{itemize}

\section{Related Work}
In this section, we briefly review the recent developments in the field of LiDAR-based place recognition. For a more comprehensive overview, the readers may refer to \cite{survey}. According to the representations used for feature extraction, we classify the current Lidar-based place recognition methods into two categories, \emph{i.e.,} the methods that utilize 3D points and the methods that use projection images as intermediate representations.

\textbf{Place recognition based on 3D points.} PointNetVLAD \cite{angelina2018pointnetvlad} leverages PointNet \cite{qi2017pointnet} to project each point into a higher dimension feature, and then uses NetVLAD \cite{arandjelovic2016netvlad} to generate global features. To take advantage of more contextual information,  {PCAN \cite{zhang2019pcan} introduces the point contextual attention network that learns attentions to the task-relevant features.} Both PointNetVLAD and PCAN cannot capture local geometric structures due to the independent treatment for each point. Thus, the following methods focus on extracting more discriminative local features considering the neighborhood information. LPD-Net \cite{liu2019lpd} adopts an adaptive local feature module to extract the handcrafted features and uses a graph-based neighborhood aggregation module to discover the spatial distribution of local features. EPC-Net \cite{epcnet} improves LPD-Net by using a proxy point convolutional neural network. DH3D \cite{du2020dh3d} designs a 3D local feature encoder to learn more distinct local descriptors, and SOE-Net \cite{soe} introduces a point orientation encoding (PointOE) module. Minkloc3D \cite{mickloc3d,mickloc3dv2} uses sparse 3D convolutions in local areas and achieves state-of-the-art performance on the benchmark dataset. Recently, some works including SVT-Net \cite{svtnet}, TransLoc3D \cite{transloc3d}, NDT-Transformer \cite{ndtformer}, and PPT-Net \cite{pptnet} leverage the transformer-based attention mechanism \cite{attention} to boost place recognition performance. However, it was shown that MinkLoc3D outperforms these transformer-based methods with fewer parameters.

\textbf{Place recognition based on projection images.} Steder \emph{et al.} \cite{2011Place} extract handcrafted local features from range images of point clouds and perform place recognition by local feature matching. Kim \emph{et al.} \cite{kim2018scan} project the point cloud into a bearing-angle image and propose the scan context descriptor. They further introduce the concept of scan context image (SCI) \cite{kim20191} and achieve place recognition by classifying the SCIs using a convolutional network. OverlapNet \cite{chen2020overlapnet} uses the overlap of range images to determine whether two point clouds are at the same place and uses a siamese network to estimate the overlap. OverlapTransformer \cite{OverlapTransformer} further uses a transformer architecture to learn rotation-invariant global features. CVTNet \cite{ma2023cvtnet} combines range images and BEV images to perform matching. It transforms BEV images into a format similar to range images to achieve rotation-invariance. Different from the aforementioned methods based on the image representations that are built under polar or polar-like projections, BVMatch \cite{bvmatch2021} projects point clouds into BEV images and extracts handcrafted BVFT features from the images. It then uses the bag-of-words model \cite{2012Bags} to generate global features. However, it is shown that BVMatch cannot generalize well to unseen environments \cite{bvmatch2021}. Different from BVMatch, we extract rotation-equivariant local features using group convolution \cite{groupconv} and generate global features by NetVLAD \cite{arandjelovic2016netvlad}. Thanks to the network design, our method can generalize to different scenes while keeping high recall rates.

\begin{figure*}[htp]
	\begin{center} 
		\includegraphics [width=5.72in]{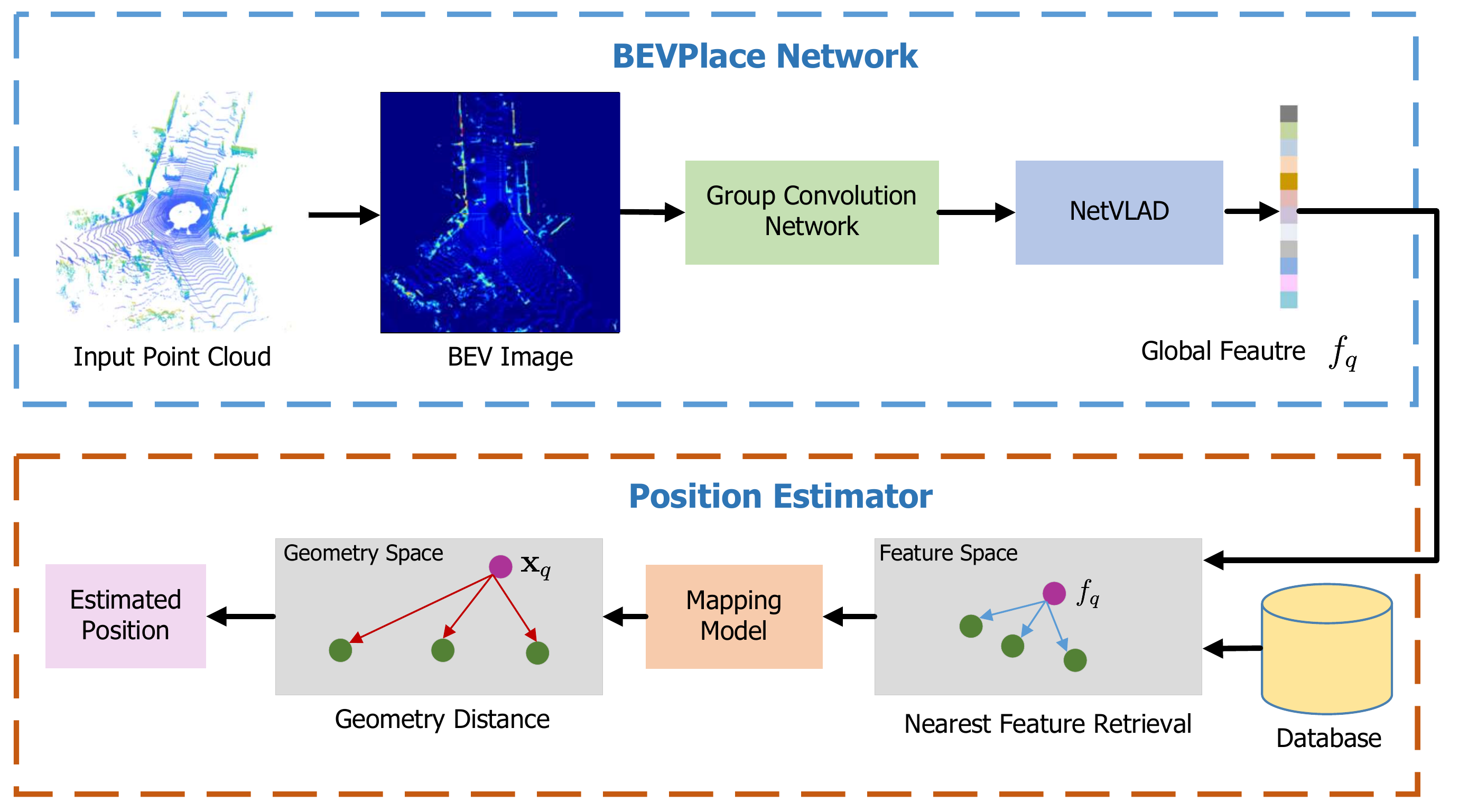}  
		\caption{Two modules of our method. In the BEVPlace network, we project point clouds into BEV images and extract rotation-invariant global features. In the position estimator module, we recover geometry distances from feature space and estimate positions of query point clouds.}
		\label{fig: method}
		\vspace{-3mm}
	\end{center} 
\end{figure*}

\section{Preliminaries}
Let $\textbf{m}_i$ be the point cloud collected by a sensor at the pose $\mathbf{T}_i=(\mathbf{R}_i,\mathbf{t}_i)$, where $\mathbf{R}_i$ is the rotation matrix and $\mathbf{t}_i$ is the position. The database formed by $n$ point clouds and their associated poses could be denoted as $\mathcal{M}=\{(\mathbf{m}_i,\mathbf{T}_i)\}_{i=1,2,...,n}$. Given a query point cloud $\textbf{m}_q$, place recognition aims at finding its most structurally similar point cloud from the pre-built database $\mathcal{M}$. In the problem of LiDAR-based place recognition, two point clouds are usually regarded as structurally similar if they are collected at geometry close places. Towards this goal, we design a network $f(\cdot)$ to map the point cloud to a distinct compact global feature vector such that $\Vert f(\mathbf{m}_q)-f(\mathbf{m}_i)\Vert_2<\Vert f(\mathbf{m}_q)-f(\mathbf{m}_j)\Vert_2$ if $\mathbf{m}_q$ is structurally similar to $\mathbf{m}_i$ but dissimilar to  $\mathbf{m}_j$.  Based on the network $f$, we perform place retrieval by finding the point cloud with the minimum feature distance to the query point cloud. 

In this work, we train our network based on BEV images of point clouds. In addition to place retrieval, we develop an extended usage that estimates the positions of the query point clouds. 

\section{Method}
Our method is formed by two modules as illustrated in Fig. \ref{fig: method}. In the BEVPlace network, we project the query point cloud into the BEV image. Then, we extract a rotation-invariant global feature through a group convolution network and NetVLAD \cite{arandjelovic2016netvlad}. In the position estimator, we retrieve the closest feature of the global feature from a pre-built database. We recover the geometry distance between the query and the matched point clouds based on a mapping model. The position of the query is estimated based on the recovered distances.

\subsection{BEVPlace Network}
In road scenes, a LiDAR sensor on a car or a robot can only move on the ground plane. Since we generate BEV images by projecting point clouds into the ground plane, the view change of the sensor will result in a rotation transformation on the image. To achieve robust place recognition, we aim at designing a network $f$ to extract rotation-invariant features from BEV images. Denoting the rotation transformation $\textbf{R}\in SO(2)$ on the BEV image $\textbf{I}$ as $\textbf{R}\circ \textbf{I}$, the rotation invariance of $f$ can be represented as 
\begin{equation}
	\label{eq: rotation_invriance}
	\begin{aligned}
		f(\textbf{R}\circ \textbf{I}) = f(\textbf{I}).
	\end{aligned}
\end{equation}
A straightforward approach to achieve such invariance is to train a network with data augmentation \cite{tipooling}. However, data augmentation usually requires that the network has a larger group of parameters to learn the rotations and may not generalize to the combination of rotations and scenes not occurring in the training set. In this work, we use the cascading of a group convolution network and NetVLAD to achieve rotation invariance. Our BEVPlace has strong generalization ability since the network is designed inherently invariant to rotations.

\textbf{Bird's Eye View Image Generation.} We follow BVMatch \cite{bvmatch2021} and use the point density to construct images. We discretize the ground space into uniform grids with a grid size of 0.4 m. For a point cloud $\mathbf{m}$, we compute the number of points in each grid and use the normalized point density as the pixel intensity of the BEV image $\mathbf{I}$.

\textbf{Group Convolution Network.} Group convolution treats the feature map as functions of the corresponding symmetry-group \cite{groupconv}. Considering the 2D rotation group $SO(2)$, applying group convolution $f_{gc}$ on a BEV image $\textbf{I}$ results in rotation-equivariant features, which can be written as 
\begin{equation}
	\label{eq: rotation_equivariance}
	\begin{aligned}
		f_{gc}(\textbf{R}\circ \textbf{I}) = \textbf{R}'\circ f_{gc}(\textbf{I}).
	\end{aligned}
\end{equation}
That is, transforming the input $\textbf{I}$ by a rotation transformation $\textbf{R}$ and then passing it through the mapping $f_{gc}$ should give the same result as first mapping $\textbf{I}$ through $f_{gc}$ and then transforming the feature with $\textbf{R}'\in SO(2)$. Usually, $f$ is designed such that $\textbf{R}'=\textbf{R}$.

Group convolution has been well-developed for a few years, and there are some mature group convolution designs \cite{groupconv, gift, e2cnn}. We implemented our network based on GIFT \cite{gift}. GIFT is originally designed for image matching and can produce distinct local features. Our main modification to GIFT is to remove the scale features since there is no scale difference between BEV images. More details of our network implementation are appended in the supplementary materials. 

\textbf{Rotation invariant global features.} According to Eq. \ref{eq: rotation_equivariance}, the contents of the feature map of group convolution keep the same for rotated images and are only transformed by a rotation $\mathbf{R}'$. Thus, we can use a global pooling operation to extract rotation-invariant global features. To capture more information about the statistics of local features, we use NetVLAD \cite{arandjelovic2016netvlad} for feature aggregation. We achieve rotation invariance by cascading the group convolution network and NetVLAD, which is,
\begin{equation}
	\label{eq: rotation_invariance}
	\begin{aligned}
		\mathrm{NetVLAD}\left(f_{gc}(\textbf{R}\circ \textbf{I})\right) &=	\mathrm{NetVLAD}\left(\textbf{R}'\circ f_{gc}(\textbf{I})\right) \\&= \mathrm{NetVLAD}\left(f_{gc}(\textbf{I})\right).
	\end{aligned}
\end{equation}



%

%
\textbf{Loss function.} There are some loss functions \cite{angelina2018pointnetvlad,soe} for LiDAR-based place recognition problem. In this work, we train our network with the simple commonly used lazy triplet loss \cite{angelina2018pointnetvlad}, formulated as 
\begin{equation}
	\label{eq: loss}
	\begin{aligned}
		\mathcal{L} = \max_j([m+\delta_\text{pos}-\delta_{\text{neg}j}]_+),
	\end{aligned}
\end{equation}
where $[. . .]_+$ denotes the hinge loss, $m$ is the margin, $\delta_\text{pos}$ is the feature distance between an anchor point cloud $\mathbf{m}_a$ and its structurally similar (``positive") point cloud, $\delta_{\text{neg}j}$ is the feature distance between $\mathbf{m}_a$ and its structurally dissimilar (``negative") point cloud. We follow the training strategy in \cite{angelina2018pointnetvlad,liu2019lpd,soe} and regard two point clouds are structurally similar if their geometry distance is less than $\epsilon$ meters.

\subsection{Position Estimator}
The lazy triplet loss forces the network to learn a mapping that preserves the adjacency of point clouds in the geometry space. Although there isn't an explicit mapping function that reveals the relationship between the feature space and the geometry space, we observe that the distance of global features and the geometry distance of point clouds are inherently correlated. Based on this property, we recover the geometry distance between the query and the match and then use it for position estimation.

\begin{figure}[htbp]
	\begin{center} 
		\includegraphics [width=3.3in]{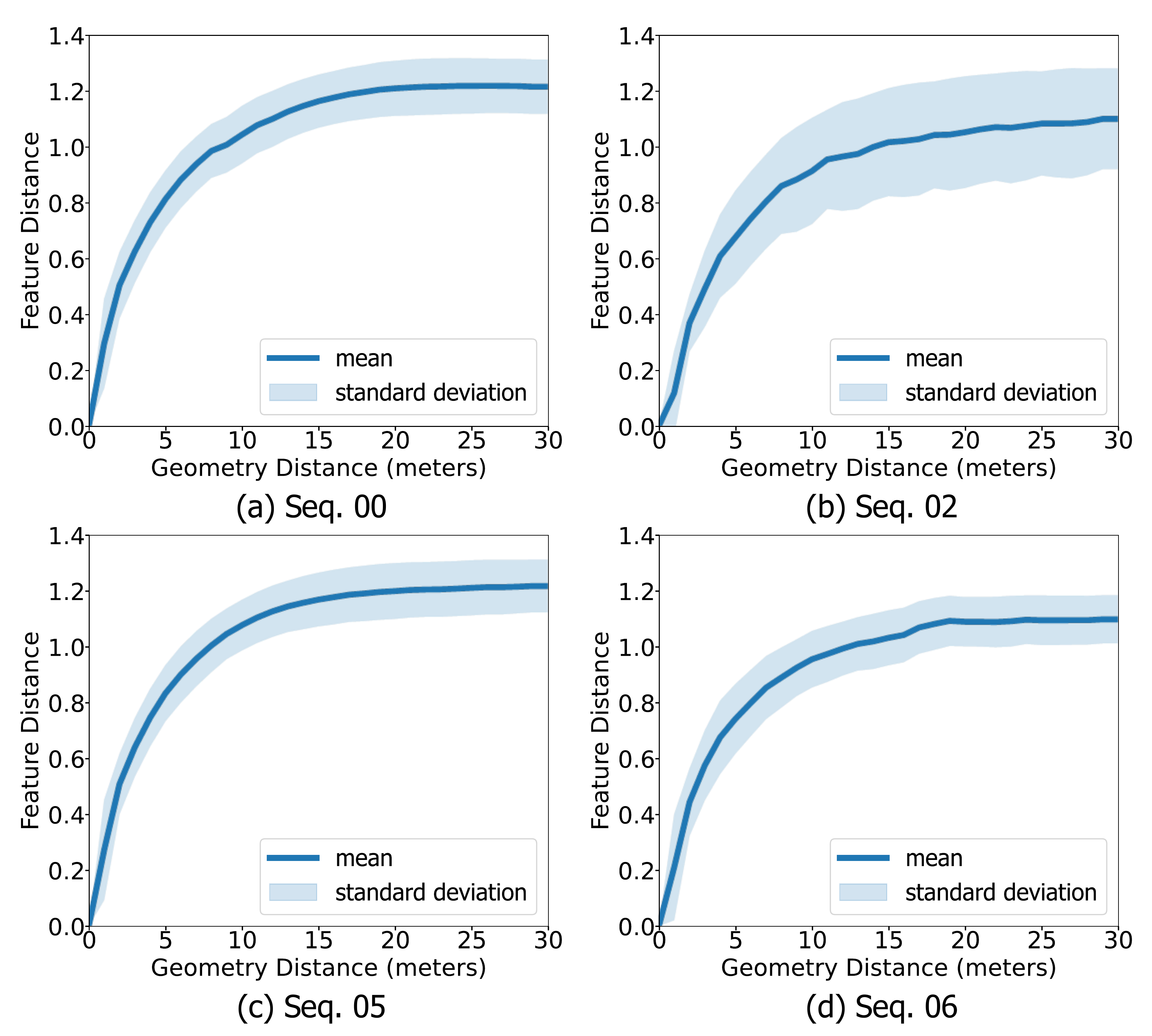}  
		\caption{ Geometry distance and feature distance relationship of the point clouds in different sequences of the KITTI dataset.}
		\label{fig: bev_distance}
		\vspace{-3mm}
	\end{center} 
\end{figure}

\textbf{Statistical correlation between the feature and geometry distances}. To reveal the relationship between the feature space and the geometry space, we train our method on the sequence ``\textit{00}'' of the KITTI dataset \cite{kitti}. We then plot the feature distances and the geometry distances of all point cloud pairs in different sequences of the dataset. As shown in Fig. \ref{fig: bev_distance}, for all the sequences, the feature distance approximately monotonically increases with the geometry distance and saturates when the point clouds are far away from each other. This phenomenon is intuitive since two point clouds are more similar if they are geometry closer, and consequently the feature distance is smaller. It can be seen that the mean curve and the standard deviation differ in different sequences since sequences are collected in diverse scenes. Despite this, the mean curves have similar shapes and can be depicted using a function based on the generalized Gaussian kernel \cite{ggd}, which is 
\begin{equation}
	\label{eq: ggd}
	\begin{aligned}
		\Vert f(\textbf{m}_i-f(\textbf{m}_j)\Vert_2 = \alpha\left(1-\exp({-\frac{\Vert\textbf{t}_i-\textbf{t}_j\Vert_2^\gamma}{\beta}})\right),
	\end{aligned}
\end{equation}
where $\alpha$ is the max feature distance, $\gamma$ and $\beta$ control the curve shape. 

\textbf{Mapping Model}. The mapping function above inspires us to recover the geometry distance from the feature distance, and further estimate positions of the query point clouds. However, this mapping relationship may differ slightly in local areas due to the appearance changes of point clouds. For more accurate geometry distance recovery, we build a mapping function for each point cloud $\mathbf{m}_i$ in the database $\mathcal{M}$. Specifically, we compute its feature and geometry distances to all the other point clouds in $\mathcal{M}$. We then fit the curve with Eq. \ref{eq: ggd} and compute the parameters $\alpha_i$, $\beta_i$, and $\gamma_i$. After this, we can recover the geometry distance of a query point cloud $\mathbf{m}_q$ to $\mathbf{m}_i$ according to Eq. \ref{eq: ggd}, that is
\begin{equation}
	\label{eq: distance_mapping}
	\begin{aligned}
		\Vert\mathbf{t}_q-\mathbf{t}_i\Vert_2 =\left(-\beta_i\log(1-\frac{\Vert f(\textbf{m}_q)-f(\textbf{m}_i)\Vert_2}{\alpha_i})\right)^{\frac{1}{\gamma_i}}.
	\end{aligned}
\end{equation}

\textbf{Position Recovery}. Since the positions of the point clouds in the database are given, we can compute the position of the query point cloud $\textbf{m}_q$ if we know its geometry distances to at least three reference point clouds. To this end, we first follow the place recognition procedure and find the most similar point cloud $\textbf{m}_r$ of $\textbf{m}_q$.  We choose the reference point clouds as $\textbf{m}_r$ and the point clouds that are less than $\epsilon$ meters away from $\textbf{m}_r$. Denoting   $\Omega = \{k\big|\Vert \mathbf{t}_r-\mathbf{t}_k\Vert_2<\epsilon\}$ and $d_k$ as the recovered geometry distance between $\textbf{m}_q$ and $\textbf{m}_k$, the position of $\textbf{m}_q$ can be easily computed by solving the following minimization problem,
\begin{equation}
	\label{eq: ggd_model}
	\begin{aligned}
		\mathbf{t}_q = \arg\min_{\mathbf{t}}\sum_{k\in\Omega}\left(\Vert\mathbf{t}-\mathbf{t}_k\Vert_2-d_k\right)^2.\\
	\end{aligned}
\end{equation}

\begin{figure}[!htp]
	\begin{center} 
		\includegraphics [width=3.3in]{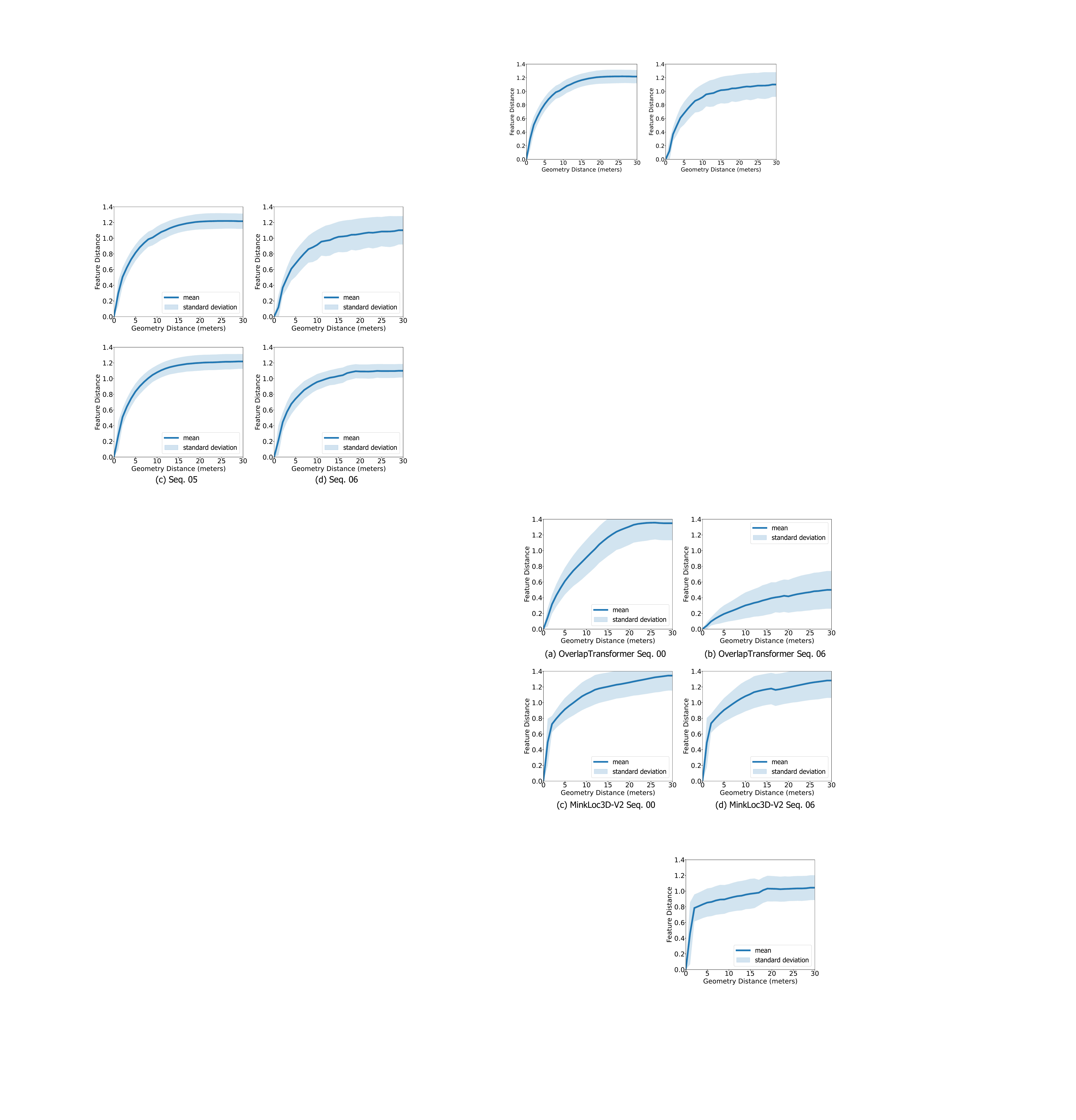}  
		\caption{ Geometry distance and feature distance relationship of the point clouds in the sequence ``\textit{00}'' and ``\textit{06}'' of KITTI of two methods.}
		\label{fig: overlap_distances}
		\vspace{-5mm}
	\end{center} 
\end{figure}
\textbf{Discussion}. In fact, the monotonicity of the mapping from feature distance to the geometry distance also holds for other methods. Fig. \ref{fig: overlap_distances} plots the relationship between the feature and the geometry spaces of two state-of-the-art methods Minkloc3D-V2 \cite{mickloc3d} and OverlapTransformer \cite{OverlapTransformer} on the sequence ``\textit{00}'' and ``\textit{06}'' of KITTI. Although the mappings of the methods have quite different shapes, they all can be approximately depicted by Eq. \ref{eq: ggd} with specific parameters and thus positions can also be estimated based on the mapping model. In the experiment, we will compare their position estimation accuracy with our method.


\section{Experiments}
We compare our method with the state-of-the-art place recognition methods including Scan Context \cite{kim2018scan}, BVMatch \cite{bvmatch2021}, PointNetVLAD \cite{angelina2018pointnetvlad}, LPD-Net \cite{liu2019lpd}, SOE-Net \cite{soe}, MinkLock3D-V2 \cite{mickloc3dv2}, and OverlapTransformer \cite{OverlapTransformer}, among which the last 5 methods are deep
learning ones. The open-sourced codes of compared methods are used for evaluation. For our method, we set the triplet margin $m=0.3$, and the number of clusters of NetVLAD as 64. In the training phase, we choose 1 positive point cloud and 10 negative point clouds in calculating loss functions. 

We test the methods in terms of place retrieval with metrics of recall at Top-1 and recall at Top-\%1. For a more comprehensive evaluation, we also compare the loop closure detection performance with the metric of Precision-Recall (PR) curve. In addition, we test the position estimation accuracy using the absolute translation error (ATE).


\subsection{Datasets}
We conduct experiments on three large-scale public datasets, i.e. the KITTI dataset \cite{kitti}, the ALITA dataset \cite{alita}, and the benchmark dataset \cite{angelina2018pointnetvlad}.

\textbf{KITTI dataset} contains a large number of point cloud data collected by a Velodyne 64-beam LiDAR under low viewpoint variation. We select the sequences ``\textit{00}'', ``\textit{02}'', ``\textit{05}'', and ``\textit{06}'' under the Odometry subset for evaluation since these sequences contain large revisited areas. We split the point clouds of each sequence into database frames and query frames for place retrieval. The partition of each sequence is summarized in Table \ref{tab: kitti_partition}. For our method, we crop each point cloud with a [$-20 $ m, $ 20 $ m] cubic window and downsample it into 4096 points. We then generate BEV images from the downsampled point clouds.  For PointNetVLAD, LPD-Net, MinkLock3D-V2, we normalize the point values  to fit their input. For OverlapTransformer, we use full point clouds since its performance is sensitive to the point density. 
\begin{table}[ht]\footnotesize
	\centering
	\caption{Dataset Partition of the KITTI dataset.} 
	\vspace{1mm}
	\begin{tabular}{lccccc}
		\toprule
		Sequence                   & 00     & 02 & 05 & 06               \\ 
		\midrule
		Database         & 0-3000       & 0-3400  & 0-1000  & 0-600            \\
		Query         & 3200-4650       & 3600-4661 & 1200-2751 & 800-1100            \\
		\bottomrule
	\end{tabular}
	\label{tab: kitti_partition}
\end{table}

\textbf{ALITA dataset} is a dataset for long-term place recognition in large-scale environments. It contains point cloud data of campus and city scenes under different illuminations and viewpoints. In this work, we use its subset released in the General Place Recognition Competition \footnote{https://www.aicrowd.com/challenges/icra2022-general-place-recognition-city-scale-ugv-localization}. We evaluate the generalization ability of the methods on its validation set and its test set. Note that the evaluation result of the test set is automatically calculated by the server once we upload the global features to the website. The point clouds of the dataset have been cropped into a [$-20 $ m, $ 20 $ m] cubic window and downsampled into 4096 points. Similar to the process in the KITTI dataset, we generate BEV images with the downsampled point clouds for our method. We normalize the points to fit the input of PointNetVLAD, LPD-Net, and MinkLock3D-V2. We do not evaluate OverlapTransformer on this dataset as it cannot adapt to such sparse point clouds.

\textbf{Benchmark dataset} is broadly used by the recent place recognition method based on unordered points. It is a dataset set consisting of four scenarios: an outdoor dataset Oxford RobotCar, three in-house datasets of a university sector (U.S.), a residential area (R.A.), and a business district (B.D.). It provides normalized point clouds of 4096 points, which can be directly used by PointNetVLAD, LPD-Net, and MinkLock3D-V2. For our method, we multiply the point values by 20 and then project the point cloud into the BEV image for feature extraction. Note that the recovered point cloud is not of the actual scale since we do not know the exact coordinate range of the point clouds. In spite of this, our method can adapt to such scale variation thanks to the convolution network design.

For the KITTI dataset and the ALITA dataset, we regard a retrieval as true positive if the geometry distance between the query and the match is less than $\epsilon=5$ meters. For the benchmark dataset, we set $\epsilon=25$ meters following the configurations in \cite{angelina2018pointnetvlad,liu2019lpd,soe,mickloc3dv2}.

\begin{figure*}[htp]
	\begin{center} 
		\includegraphics [width=6.8in]{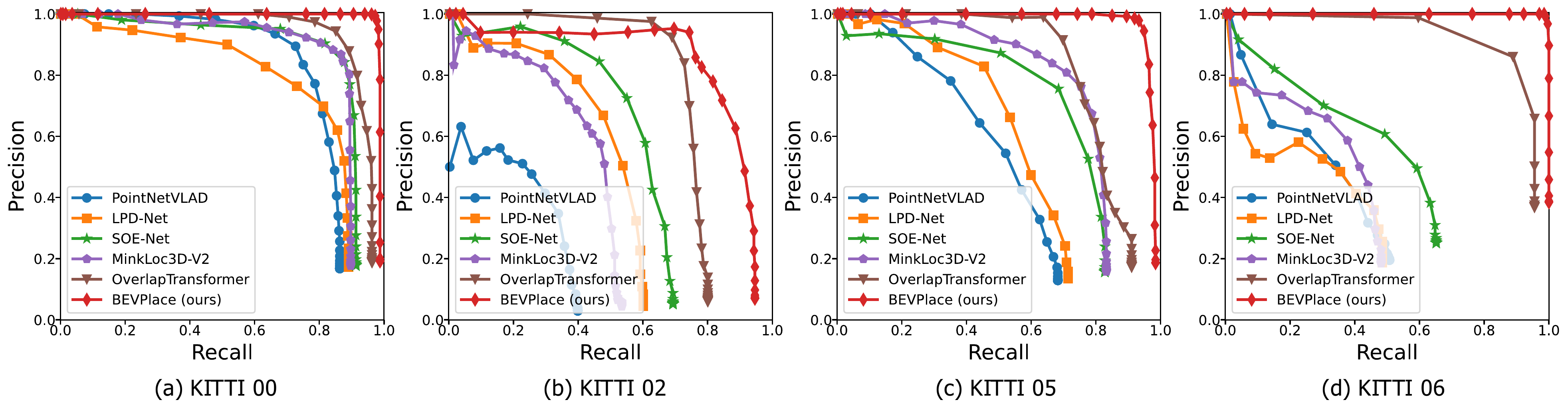}  
		\caption{Precision-recall curves for the sequences of the KITTI dataset.}
		\label{fig: pr_curve}
		\vspace{-3mm}
	\end{center} 
\end{figure*}

\subsection{Place Recognition}
We train the methods with the point clouds in the database of sequence ``\textit{00}'' of the KITTI dataset. During the training phase, we apply data augmentation by randomly rotating the point clouds around the z-axis (which is perpendicular to the ground) within the interval of $[-\pi,\pi)$.

\textbf{Ablation on the BEV representation and Group Convolution Network.} To validate the claim that a NetVLAD network based on BEV images can achieve comparable performance to the SOTA methods, we implemented two NetVLAD networks with ResNet34 and ResNet18 as backbones, respectively. To explore the influence of rotation-invariance designs to the unordered points based methods, we additionally implement a network VN-PointNetVLAD by replacing the backbone of PointNetVLAD with Vector Nureon \cite{deng2021vector}.  We observe the results shown in Table. \ref{tab: ablation} and find that, 
\begin{itemize}
	 \item Without any delicate design, both NetVLAD networks achieves comparable recalls to OverlapTransformer and outperforms the unordered points based methods. While BVMatch also leverages BEV images, it generates global descriptors using the bag-of-words model \cite{2012Bags}, which leads to inferior performance compared to the deep learning-based BEV methods. 
	 \item Our BEVPlace achieves higher recalls than the NetVLAD networks, validating the significance of the rotation invariance design. 
	 \item Among the rotation-invariant methods based on different representations, i.e. VN-PointNetVLAD based on unordered points, OverlapTransformer based on range images, and BEVPlace based on BEV images, our BEVPlace achieves higher recall and better generalization ability than the non-BEV networks, further indicating the importance of the BEV representation in place recognition.
\end{itemize}

\begin{table}[htp]\footnotesize
	\centering
	%
	\caption{Recall at Top-1 on the KITTI dataset. * denotes that method is designed rotation-invariant.} 
	\vspace{1mm}
	\begin{tabular}{lccccccccc}
		\toprule
		Sequence & 00 & 02 & 05 & 06 & Mean  \\
		\midrule
		Scan Context \cite{kim2018scan} & 89.7 & 73.9 & 77.0 & 86.7 & 81.8 \\
		PointNetVLAD\cite{angelina2018pointnetvlad}  &91.6&62.3&76.9&77.8 & 77.2\\
		LPD-Net     \cite{liu2019lpd}         &95.7&72.3&83.6&82.2 & 83.5\\
		SOE-Net     \cite{soe}         &95.0&65.5&84.8&69.6& 78.7\\
		MinkLoc3D-V2  \cite{mickloc3d}       &95.9&72.3&86.4&80.4 & 83.4\\
		*BVMatch \cite{bvmatch2021}     & 93.8 & 78.2 & 90.2 & 93.8 & 89.0 \\
		*VN-PointNetVLAD \cite{angelina2018pointnetvlad,deng2021vector} &94.3&66.5&87.5&84.3 &82.9\\
		*OverlapTransformer\cite{OverlapTransformer}   &96.7&80.1&91.9&95.6& 91.1\\
		\midrule
		ResNet18+NetVLAD & 95.9 & 83.2 & 90.3 & 98.5 & 92.0\\
		ResNet34+NetVLAD & 96.3 & 84.1 & 92.2 & 98.5 & 92.8 \\ 
		*BEVPlace (ours)     &\textbf{99.7} &\textbf{98.1} &\textbf{99.3} &\textbf{100.0} & \textbf{99.3} \\
		\bottomrule
	\end{tabular}
	
	\label{tab: ablation}
\end{table}

\begin{table}[!htp]\footnotesize
	\centering
	\caption{Recall at Top-1 on the rotated KITTI dataset. * denotes that method is designed rotation-invariant.} 
	\vspace{1mm}
	\begin{tabular}{lccccccccc}
		\toprule
		Sequence & 00 & 02 & 05 & 06 &Mean  \\
		\midrule
		Scan Context \cite{kim2018scan} & 89.7 & 73.9 & 77.0 & 86.7 & 81.8 \\
		PointNetVLAD\cite{angelina2018pointnetvlad}  &86.1&41.0&69.7&51.5 & 62.1\\
		LPD-Net     \cite{liu2019lpd}         &89.6&61.9&72.2&48.9 & 68.2\\
		SOE-Net     \cite{soe}         &93.1&63.5&82.8&65.5 & 76.2\\
		MinkLoc3D-V2  \cite{mickloc3d}       &89.4&48.7&83.0&48.1 & 67.3\\
		*BVMatch \cite{bvmatch2021}       &   93.5 & 77.8 & 89.1 & 92.6 & 88.6 \\
		*VN-PointNetVLAD                      &   93.2 & 62.3 & 85.2  & 82.9 & 80.9 \\
		*OverlapTransformer\cite{OverlapTransformer}   &96.7&80.1&91.9&95.6 & 91.1\\
		\midrule
		ResNet18-NetVLAD   \cite{vgg}              &92.3&64.1&89.8&89.9 & 84.0\\
		ResNet34+NetVLAD   \cite{arandjelovic2016netvlad} & 93.1 & 64.2 & 90.7 & 90.4 & 84.6\\
		*BEVPlace (ours)     &\textbf{99.6} &\textbf{93.5} &\textbf{98.9} &\textbf{100.0} & \textbf{98.0}  \\
		\bottomrule
	\end{tabular}
	\label{tab: top1_kitti_rot}
\end{table}

\textbf{Robustness to view changes.} In the testing phase, we randomly rotate the point clouds of KITTI along z-axis to simulate view changes. As shown in Table \ref{tab: top1_kitti_rot}, our method shows much higher recall rates than ResNet18-NetVLAD and ResNet34+NetVLAD. It is also noted that VN-PointNetVLAD performs better than PointNetVLAD. These validate the significance of the rotation invariance design to view variations. However, VN-PointNetVLAD cannot generalize well to sequences ``\textit{00}'', ``\textit{02}'', ``\textit{05}'', and ``\textit{06}''. On the other hand, our BEVPlace shows much higher recall and better generalization ability than all the other methods.  

\textbf{Loop closure detection.}
Loop closure detection is an important application of place recognition. For a query point cloud, we accept its Top-1 match as positive if the feature distance is less than a threshold. By setting different thresholds, we compute the precision-recall curves and plot them in Fig. \ref{fig: pr_curve}. It can be seen that our method outperforms the compared methods. It is worth noting that although our method is only trained on a part of the point clouds of the sequence $00$, it generalizes much better to the other sequences than the other methods. We believe our method can be deployed by LiDAR SLAM systems \cite{loam} and help globally consistent mapping building. 

\begin{table}[htp]\footnotesize
	\centering
	\caption{Recall rates on the ALITA dataset.} 
	\vspace{1mm}
	\begin{tabular}{lcccccccc}
		\toprule
		& \multicolumn{2}{c}{\dlmu[1.8cm]{Val Set}} & \multicolumn{2}{c}{\dlmu[1.8cm]{Test set}}  \\
		& @1     & @1\%  & @1     & @1\%     \\ 
		\midrule
		PointNetVLAD\cite{angelina2018pointnetvlad}  &42.3&55.4&39.8&-\\
		LPD-Net     \cite{liu2019lpd}         &51.2&72.7&49.6&-\\
		SOE-Net     \cite{soe}         &66.6&92.8&59.5&-\\
		MinkLoc3D-V2  \cite{mickloc3d}       &55.6&82.8&55.3&-\\
		\midrule
		BEVPlace (ours)     &\textbf{96.7} &\textbf{99.2} &\textbf{91.7} &-  \\
		\bottomrule
	\end{tabular}
	\label{tab: recall_alita}
\end{table}

\begin{table*}[!]\footnotesize
	\centering
	\caption{Recall rates on the benchmark dataset.} 
	\vspace{1mm}
	\renewcommand\tabcolsep{2pt}
	\begin{tabular}{lcccccccc|cc}
		\toprule
		& \multicolumn{2}{c}{\dlmu[2.5cm]{Oxford}}     & \multicolumn{2}{c}{\dlmu[2.5cm]{U.S.}}       & \multicolumn{2}{c}{\dlmu[2.5cm]{R.A.}}       & \multicolumn{2}{c}{\dlmu[2.5cm]{B.D}}        &
		\multicolumn{2}{|c}{Mean}   \\ 
		& AR@1 & AR@1\% & AR@1 & AR@1\% & AR@1 & AR@1\% & AR@1 & AR@1\% & AR@1 & AR@1\%  \\
		\midrule
		PointNetVLAD \cite{angelina2018pointnetvlad}        & 62.8 & 80.3  & 63.2 & 72.6 & 56.1 & 60.3 & 57.2   & 65.3      & 59.8       & 69.6   \\ 
		LPD-Net \cite{liu2019lpd}   & 86.3       & 94.9 & 87.0 & 96.0 & 83.1  & 90.5  & 82.5 & 89.1 & 84.7 & 92.6    \\ 
		NDT-Transformer \cite{ndtformer}      & 93.8       & 97.7          & -          & -      & -       & -       & -       & -     & -       & -  \\ 
		PPT-Net    \cite{pptnet}           & 93.5       & 98.1       & 90.1       & 97.5   & 84.1       & 93.3       & 84.6       & 90.0    & 88.1       & 94.7   \\ 
		SVT-Net    \cite{svtnet}             & 93.7       & 97.8       & 90.1       & 96.5   & 84.3       & 92.7       & 85.5       & 90.7    & 88.4       & 94.4   \\ 
		TransLoc3D \cite{transloc3d}         & 95.0       & 98.5          & -          & -      & -       & -       & -       & -    & -       & -   \\ 
		MinkLoc3Dv2  \cite{mickloc3dv2}            & 96.3       & 98.9       & 90.9       & 96.7   & 86.5       & 93.8       & 86.3       & 91.2    & 90.0       & 95.1   \\ 
		\midrule
		ResNet34+NetVLAD  \cite{mickloc3dv2}            & 95.8       &  \textbf{99.2 }     & 96.2       & 99.0   &  90.1       & \textbf{99.4}       & 94.8       & \textbf{100.0}    & 94.2       & \textbf{99.4}   \\ 
		BEVPlace (ours)  & \textbf{96.5}   & {99.0}   & \textbf{96.9}   & \textbf{99.7}   & \textbf{92.3}         & {98.7}   & \textbf{95.3}         & {99.5}   & \textbf{95.3}       & {99.2}  \\
		\bottomrule %
	\end{tabular}
	\vspace{-1mm}
	\label{tab: recall_oxford}
\end{table*}

\textbf{Generalization performance on ALITA.} We test the place recognition performance of the methods on the ALITA dataset based on the model trained on the KITTI dataset. Table \ref{tab: recall_alita} shows the recall rates on the validation set and the test set. It can be seen that our method generalizes well in ALITA. On the other hand, the recall rates of the compared methods degrade much.

%

\textbf{Performance on benchmark datasets.} Following the previous works, we train our method using only the Oxford RobotCar training dataset and test the method on the test set. The details of the dataset partition can be found in \cite{angelina2018pointnetvlad}. For a more comprehensive comparison, we also compare our method with the state-of-the-art transformer-based methods, including NDT-Transformer \cite{ndtformer}, PPT-Net \cite{pptnet}, SVT-Net \cite{svtnet}, and TransLoc3D \cite{transloc3d}. For all the compared methods, we directly use the results from their papers. Table \ref{tab: recall_oxford} shows that the ResNet34+NetVLAD network achieves comparable recall to the state-of-the-art method minklock3D-v2 and shows better generalization ability. Our BEVPlace outperforms other methods including the transformer-based ones with large margins.

\subsection{Position Estimation}
We first recovery the geometry distances between the query and the matches and then estimate the global position of point clouds. In the following, we evaluate the performance of these two  phases.

\begin{figure}[!h]
	\begin{center} 
		\includegraphics [width=1.8in]{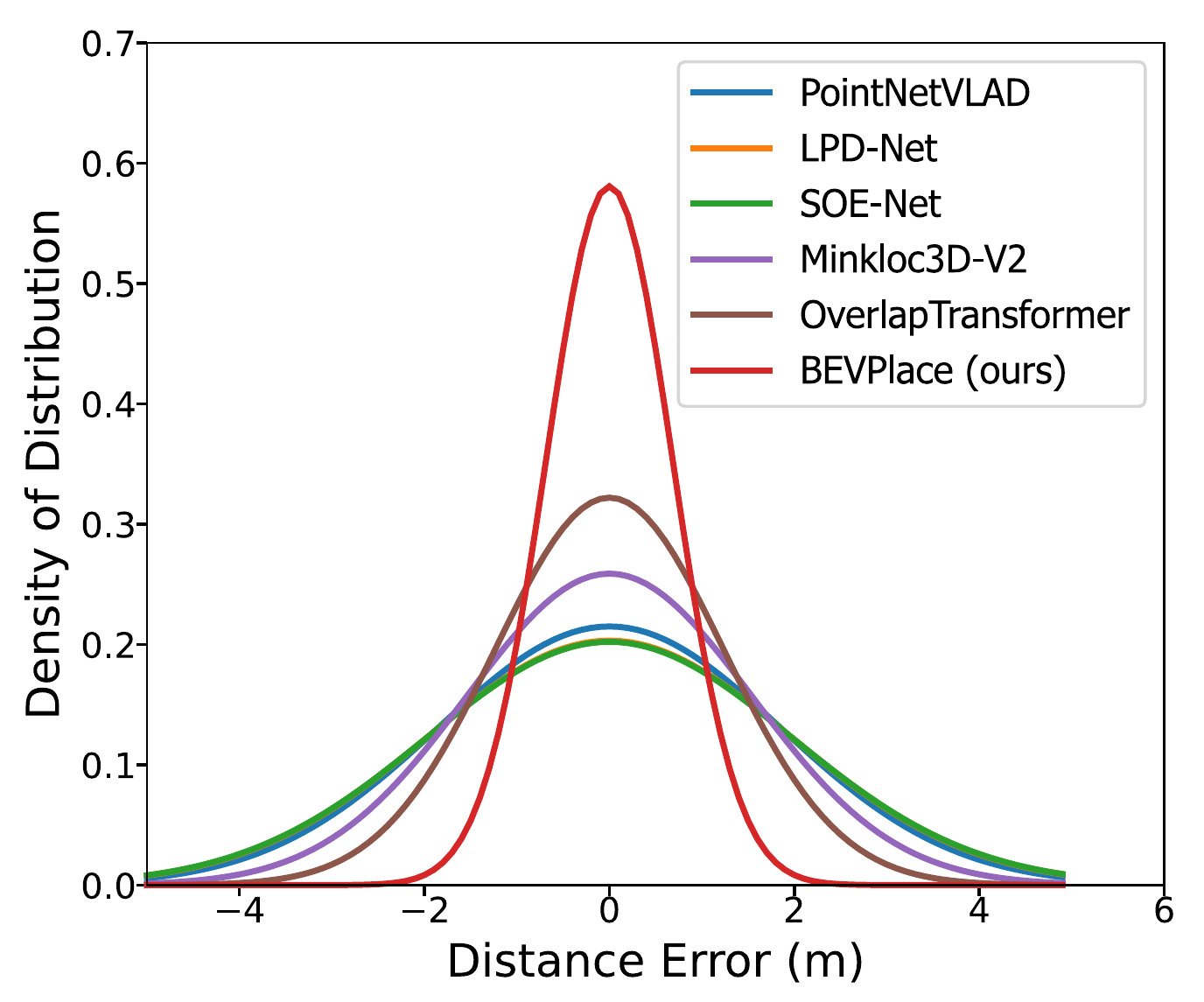}  
		\caption{ Distance estimation error distribution.}
		\label{fig: fit_error}
		\vspace{-4mm}
	\end{center} 
\end{figure}

\textbf{Accuracy of the recovery distances}. We compute the errors of the recovered distances on the sequence ``00'' of the KITTI dataset. Fig. \ref{fig: fit_error} shows the fitting distribution of the distance errors of the methods. It can be seen that our method can recover the geometry distance more accurately. This will lead to more accurate position estimation results since the estimation is based on the recovered distances.

\begin{figure*}[!ht]
	\begin{center} 
		\includegraphics [width=6in]{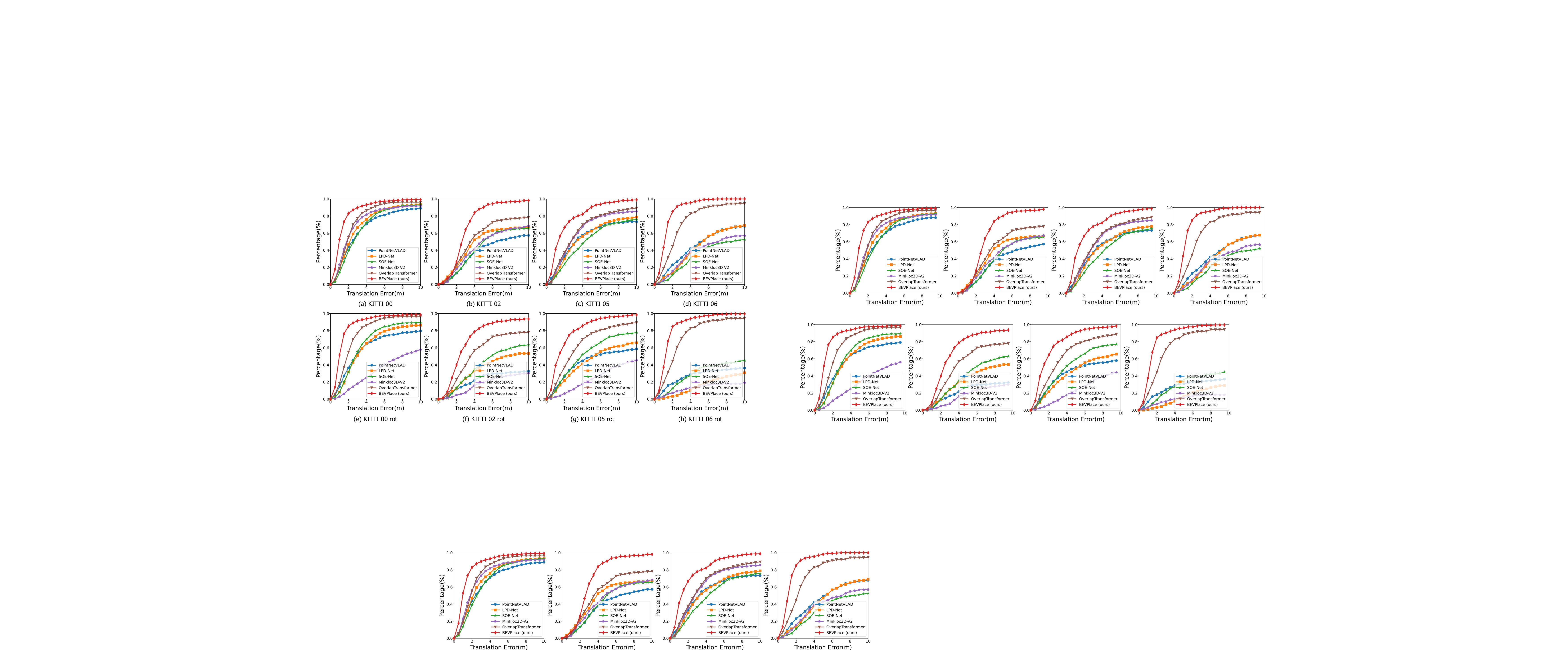}  
		\caption{ Accumulative translation error distribution on the KITTI dataset with and without rotations.}
		\label{fig: acc_error}
	\end{center} 
	\vspace{-3mm}
\end{figure*}

\textbf{Position estimation}. Fig. \ref{fig: acc_error} (a), (b), (c), and (d) show the cumulative distribution of the translation error on different sequences of the KITTI datasets. Our method and OverlapTransformer, both of which are based on projection images, achieve more accurate position estimation than the compared methods. To validate the performance under view changes, we randomly rotate the point clouds in the testing phase. Fig. \ref{fig: acc_error} (e), (f), (g), and (h) show that our method and OverlapTransformer perform well since they are designed to be rotation invariant. On the other hand, the other methods shows poor robustness and their performance degrades much.

\subsection{Running Time}
We evaluate the runtime of BEVPlace on a desktop with an RTX3090 GPU. For each query, place recognition costs about $30$ ms, and position estimation costs less than $1$ ms. Our method can realize real-time localization as most LiDAR sensors operate at 10 Hz.

\section{Conclusions}
In this work, we explore the potential of LiDAR-based place recognition using BEV images. We designed a rotation invariant network called BEVPlace based on group convolution. Thanks to the use of BEV images and the rotation invariance design, our method achieves high recall rates, strong generalization ability, and robustness to viewpoint changes, as shown in the experiments. In addition, we observe that the geometry and feature distance are correlated, and we model the correlation for position estimation. This model can adapt to other place recognition methods, but our BEVPlace gives more accurate estimation results. In our future work, we will try to encode the rotation information into global features and estimate 6-DoF pose of point clouds. 

\textbf{Acknowledgement.} This work was supported in part by the Ten Thousand Talents Program of Zhejiang Province under grant 2020R52003 and in part by the National Natural Science Foundation of China under grant 62002323.

{\small
\bibliographystyle{ieee_fullname}
\bibliography{egbib}
}

\end{document}